\documentclass{article}

\usepackage{microtype}
\usepackage{graphicx}
\usepackage{caption}
\usepackage{subcaption}
\usepackage{booktabs}

\usepackage{amsmath}
\usepackage{amssymb}
\usepackage{multirow}
\usepackage{enumitem}
\usepackage{siunitx}
\usepackage{dsfont}
\newcommand{\R}{\mathbb{R}}

\newcommand{\bz}{\textbf{z}}
\newcommand{\bZ}{\textbf{Z}}

\usepackage[accepted]{icml2020}

\usepackage{hyperref}

\icmltitlerunning{DeBayes: a Bayesian Method for Debiasing Network Embeddings}

\begin{document}
	
\twocolumn[
\icmltitle{DeBayes: a Bayesian Method for Debiasing Network Embeddings}

\begin{icmlauthorlist}
	\icmlauthor{Maarten Buyl}{ugent}
	\icmlauthor{Tijl De Bie}{ugent}
\end{icmlauthorlist}

\icmlaffiliation{ugent}{Department of Electronics and Information Systems (ELIS), IDLab, Ghent University, Ghent, Belgium}

\icmlcorrespondingauthor{Maarten Buyl}{maarten.buyl@ugent.be}
\icmlcorrespondingauthor{Tijl De Bie}{tijl.debie@ugent.be}

\icmlkeywords{DeBayes, debias, fairness, network embedding, link prediction, graph embedding, conditional network embedding, cne}

\vskip 0.3in
]

\printAffiliationsAndNotice

\begin{abstract}
As machine learning algorithms are increasingly deployed for high-impact automated decision making, ethical and increasingly also legal standards demand that they treat all individuals fairly, without discrimination based on their age, gender, race or other sensitive traits. In recent years much progress has been made on ensuring fairness and reducing bias in standard machine learning settings. Yet, for network embedding, with applications in vulnerable domains ranging from social network analysis to recommender systems, current options remain limited both in number and performance. We thus propose DeBayes: a conceptually elegant Bayesian method that is capable of learning \textit{debiased} embeddings by using a \textit{biased} prior. Our experiments show that these representations can then be used to perform link prediction that is significantly more fair in terms of popular metrics such as \textit{demographic parity} and \textit{equalized opportunity}.
\end{abstract}

\section{Introduction}\label{sec:intro}
In recent years, machine learning algorithms have increasingly taken up decision making responsibilities in the real world. However, their traditional focus on accuracy makes them prone to discriminate in ways that are morally or legally unacceptable. This development has sparked a surge of research interest in machine learning \textit{fairness} \cite{friedler2019comparative}, where the objective is to learn how to make decisions that remain accurate while avoiding to discriminate against individuals with respect to a particular set of \textit{sensitive attributes}. 

A naive attempt at preventing discrimination is to simply remove the sensitive information from the input, thereby avoiding what is seen as \textit{disparate treatment}. Yet moral standards and anti-discrimination laws are typically also concerned about \textit{disparate impact} \cite{barocas2016big}. This alternative notion of discrimination demands that outcomes do not disproportionately hurt or benefit people with specific sensitive attributes \cite{zafar2017fairness}. In machine learning, avoiding disparate impact is often difficult, since the sensitive attributes are typically not independent of other features in the data \cite{pedreshi2008discrimination}.
f
While research in fair binary classification has been plentiful, fairness has seen relatively little attention in the field of \textit{network embedding} \cite{mehrabi2019survey}. There, the goal is to find a low-dimensional representation for every node in a graph dataset, in order to perform various downstream tasks such as node classification, community detection, and link prediction \cite{cui2018survey}. The latter is concerned with estimating, given the observed network structure, the likelihood of an edge between two nodes. The field has seen tremendous interest in recent years due to the ubiquity of the graph structured data, with applications in e.g. social graphs and recommender systems \cite{hamilton2017representation}. However, these networks often represent connections between people, possibly giving rise to fairness issues. For example, friendship recommendation in social media graphs may reinforce historical biases towards certain groups \cite{stoica2018algorithmic}. A similar issue has been found in book recommendation, where algorithms may be biased towards suggesting books with male authors \cite{ekstrand2018exploring}.

In this work, our aim is to provide an elegant method to increase fairness at both the level of network embedding and at the level of its downstream task: link prediction. An algorithm that naturally allows us to do so is \textit{Conditional Network Embedding} (CNE) \cite{kang2019conditional}: a Bayesian network embedding algorithm that uses a \textit{prior distribution} to model the structural properties of the network, such as the node degrees or any block structure. Given this, CNE then searches for an embedding that optimally describes the affinities between nodes \emph{insofar these are not already captured by the prior}.

\begin{figure}[tb]
	\centering
	\begin{subfigure}[b]{0.45\linewidth}
		\includegraphics[width=\linewidth]{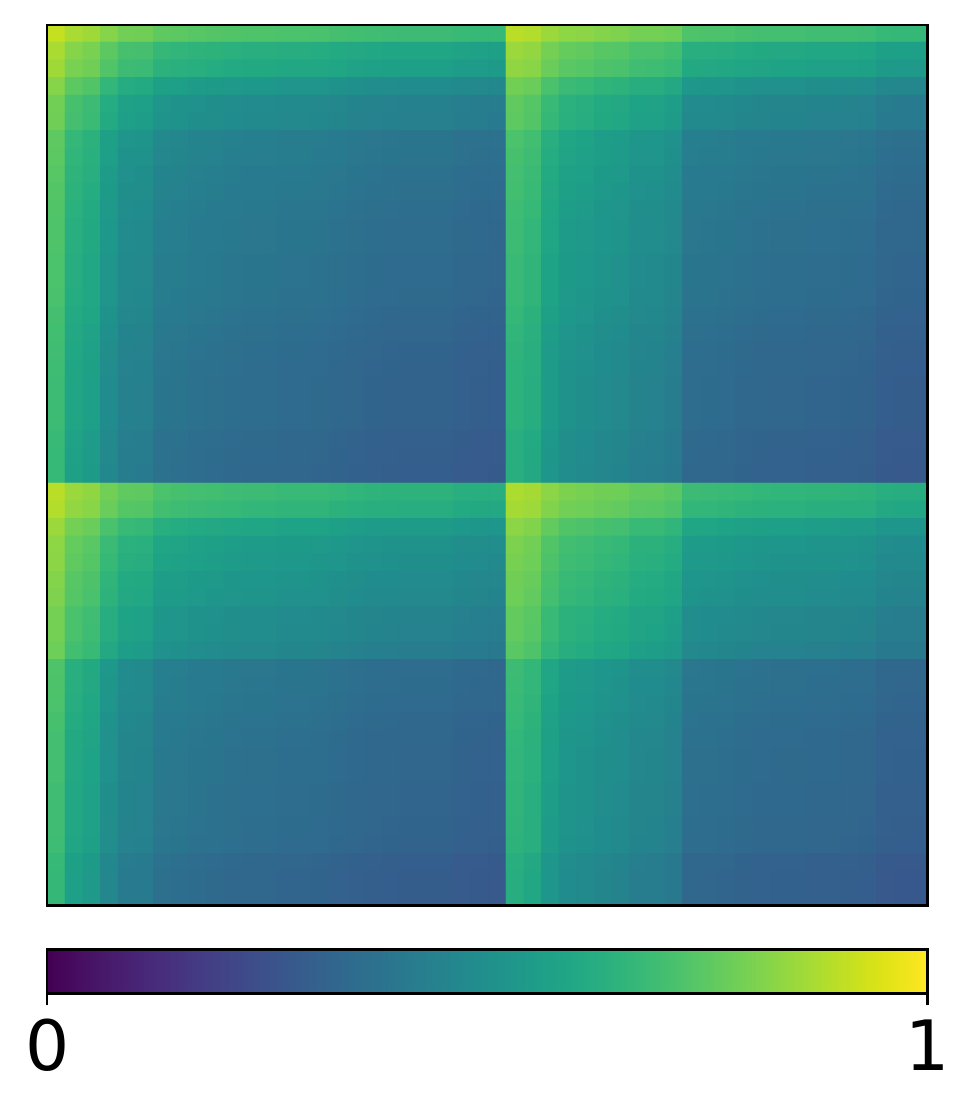}
		\caption{oblivious training prior}
		\label{subfig:obli_prior}
	\end{subfigure}
	\begin{subfigure}[b]{0.45\linewidth}
		\includegraphics[width=\linewidth]{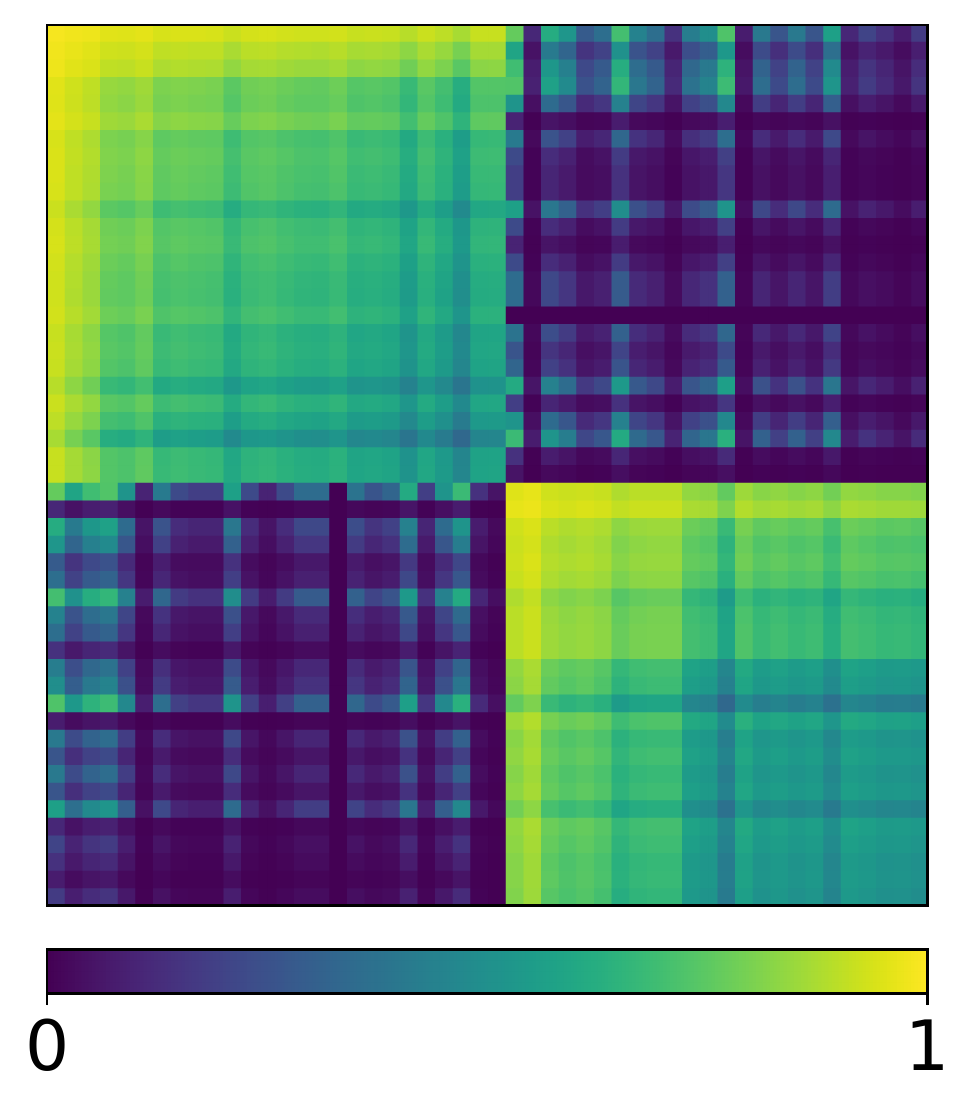}
		\caption{biased training prior}
		\label{subfig:bias_prior}
	\end{subfigure}

	\begin{subfigure}[b]{0.45\linewidth}
		\includegraphics[width=\linewidth]{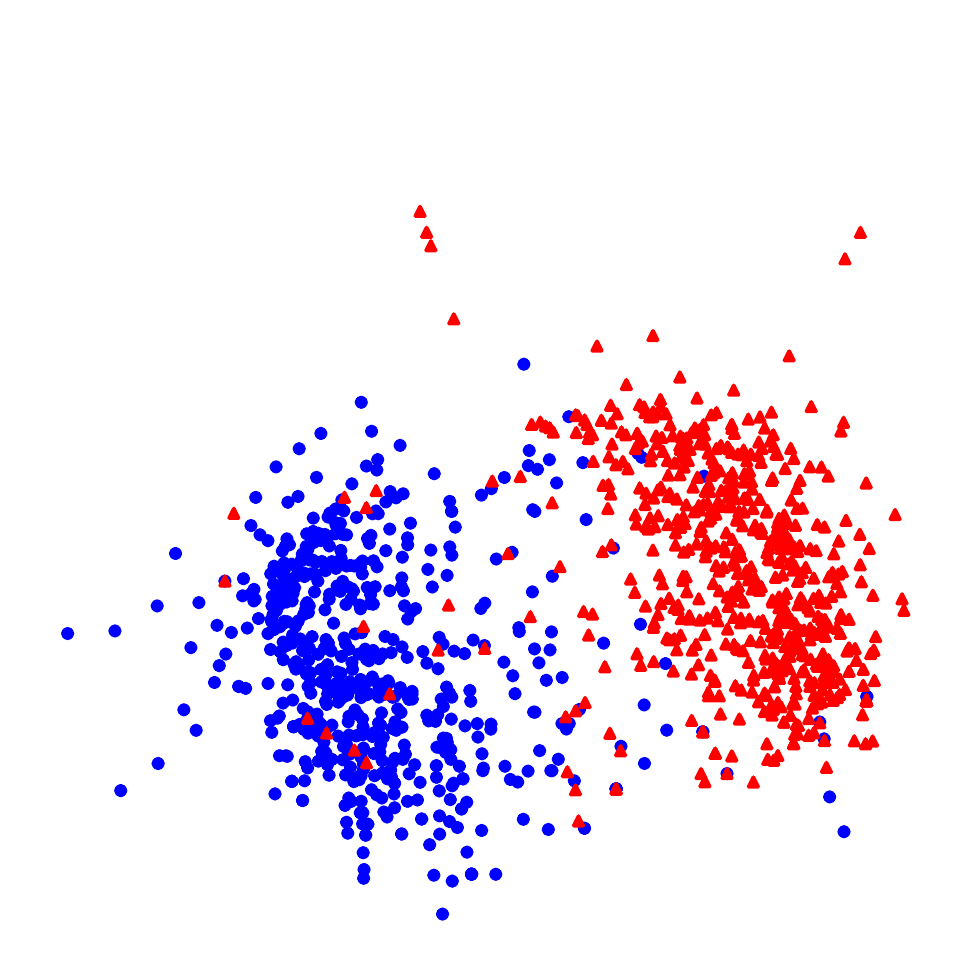}
		\caption{CNE embeddings}
	\end{subfigure}
	\begin{subfigure}[b]{0.45\linewidth}
		\includegraphics[width=\linewidth]{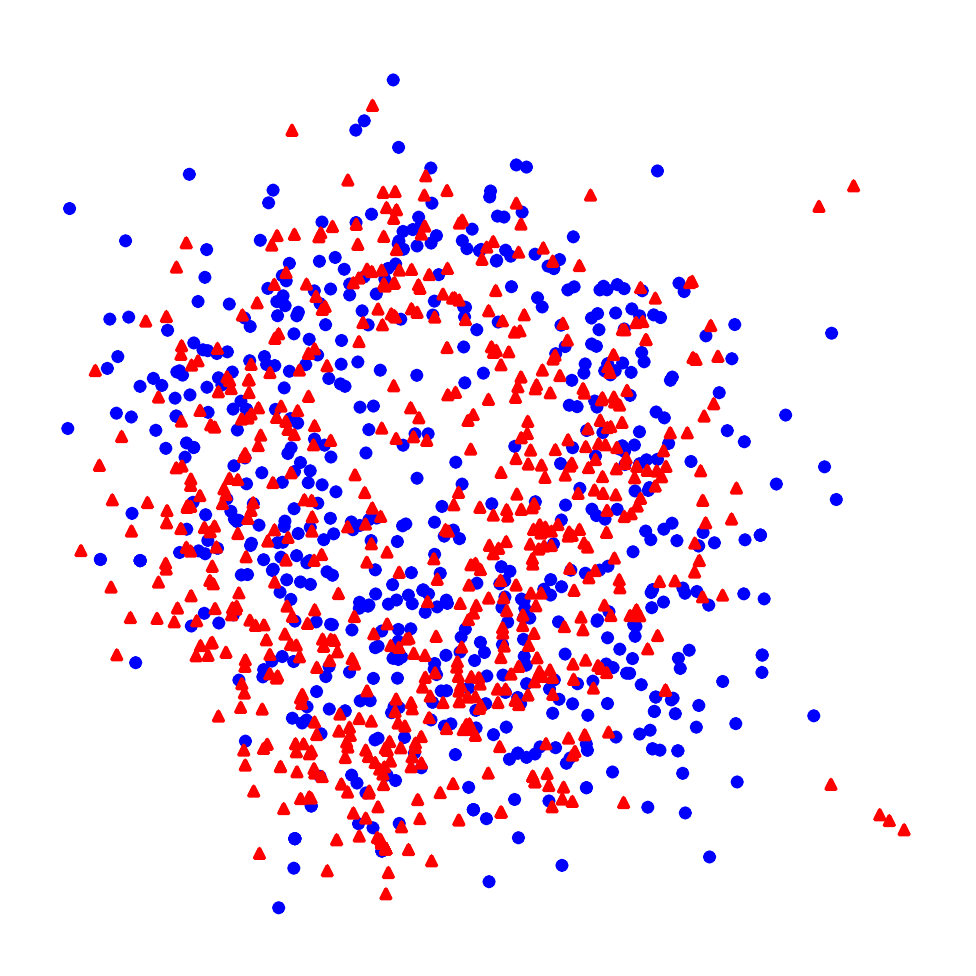}
		\caption{DeBayes embeddings}
	\end{subfigure}
	\caption{An illustration of DeBayes' main idea. The nodes represent political blogs on US politics and they have an edge if one of them hyperlinks to the other. The sensitive attribute is whether the blog author leans liberal (blue circles) or conservative (red triangles). The matrices (a, b) display, for a link between blogs $i$ and $j$, the prior probability at location $(i, j)$ as a heatmap. Only the 50 highest degree nodes are shown, grouped by political leaning and sorted by degree. The scatter plots (c,d) show the 2-dimensional embeddings that CNE learns if the respective prior is used.}
	\label{fig:debias_example}
	\vskip -0.05in
\end{figure}

\textbf{Proposed method}~~  We introduce \textit{DeBayes}, an adaptation of CNE where the sensitive information is modelled as strongly as possible in the prior distribution. Consequently, the learned embeddings no longer need to represent this information, i.e. it is debiased. The idea is illustrated with a 2-dimensional embedding of the \hyperref{http://www-personal.umich.edu/\~mejn/netdata/}{}{}{\textit{political blogs}} network \cite{adamic2005political} in Figure~\ref{fig:debias_example}. Panel (a) shows an \emph{oblivious} prior probability matrix, which takes the observed node degrees into account, but is oblivious to the sensitive attribute (political leaning). Nodes are grouped per political leaning (liberal and conservative), and sorted in order of decreasing degree within each group. Panel (c) shows the corresponding embedding found by CNE, clearly showing a separation between both groups, as it needs to represent the fact that same-leaning blogs are connected more frequently. Panels (b) and (d) on the other hand show a biased prior (edges between groups are less probable than within) and the matching CNE embedding. Here no separation is required as the prior already represents the bias.

\begin{figure}[tbp]
\begin{center}
\centerline{\includegraphics[width=0.8\columnwidth, trim=9.65cm 3.7cm 9.7cm 2.8cm, clip]{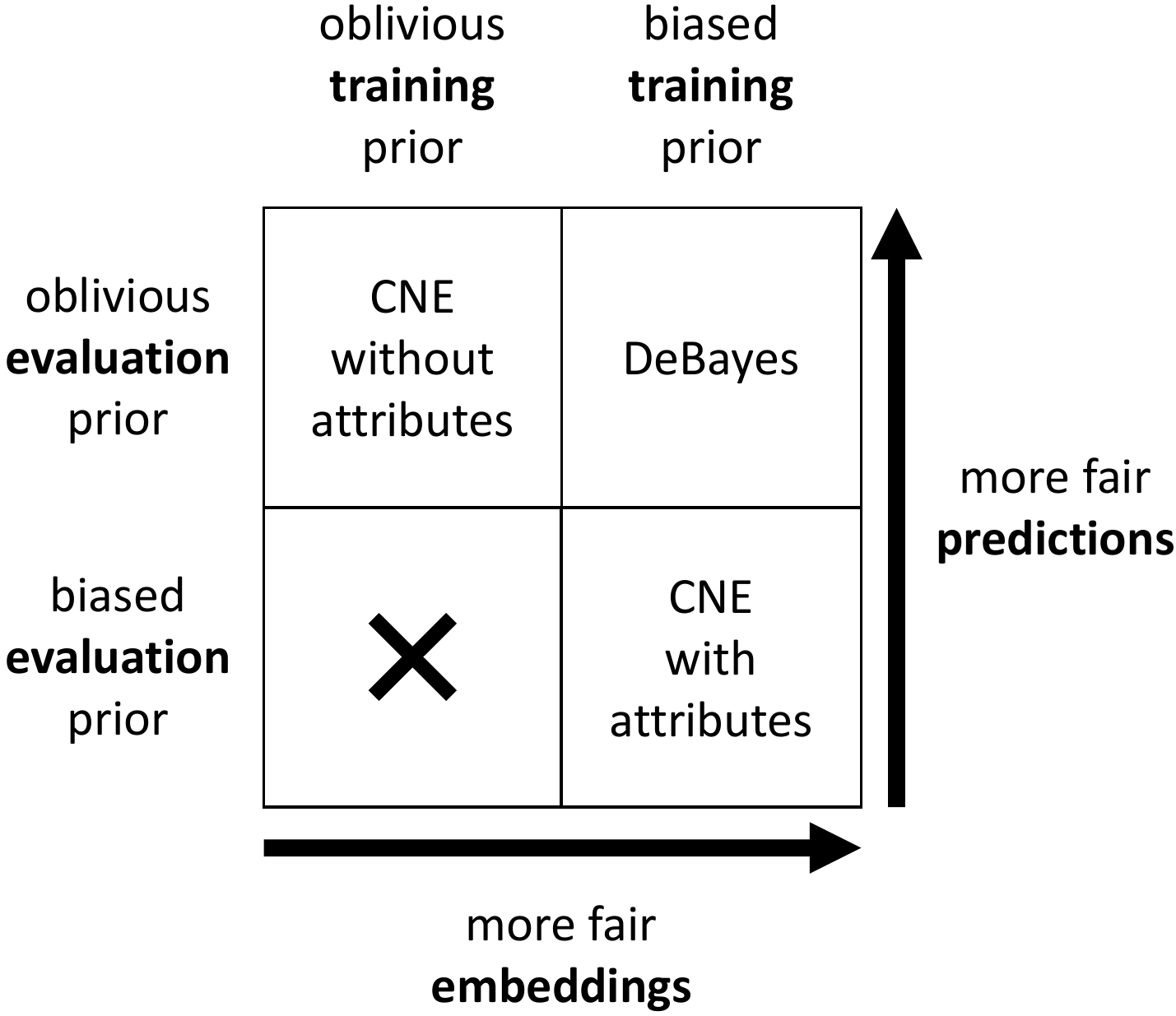}}
\caption{The results of different combinations of oblivious or biased priors at training and evaluation time. Note that the combination of `oblivious training prior' and `biased evaluation prior' is rarely desirable, as it enforces unfairness while likely decreasing prediction accuracy.}
\label{fig:priors_dia}
\end{center}
\vskip -0.25in
\end{figure}

Of course, link predictions with a CNE embedding and matching prior will discriminate regardless of the prior: if the prior does not represent the bias, then the embedding will. DeBayes' strategy, however, is to \emph{train} the CNE embedding using a biased prior, resulting in debiased embedding, but then to \emph{evaluate} the probability of links (i.e. do link prediction) in combination with the oblivious prior, as illustrated in Figure~\ref{fig:priors_dia}. We will show that this strategy reduces bias, while keeping link prediction accuracy high.

DeBayes has several conceptual advantages. First, it is simple to choose whether fairness is applied or not at evaluation time; it suffices to choose the required evaluation prior. Second, the prior is flexible enough to model several sensitive attributes, such as gender and age, at the same time. Third, imposing fairness in this way implies no additional computational cost. Finally, the concept behind DeBayes could be extended to other fields within representation learning.

\textbf{Outline}~~
The rest of the paper is organized as follows. Section~\ref{sec:related} briefly surveys related work. Section~\ref{sec:fairness_in_lp} proposes a measure of what is meant by `bias in embeddings', and also extends the well-known fairness definitions \textit{demographic parity} and \textit{equalized opportunity} to link prediction. Section~\ref{sec:method} discusses how the priors of DeBayes are constructed and how CNE is adapted to generate debiased embeddings and link predictions. Section~\ref{sec:experiments} empirically confirms DeBayes' superiority when compared with state-of-the-art baselines. Section~\ref{sec:conclusion} offers conclusions and a forward look.

\section{Related Work}\label{sec:related}
Broadly speaking, the fairness of a machine learning model can be improved in one of three stages in its pipeline \cite{friedler2019comparative}: in preprocessing, in the algorithm itself or in postprocessing. The \textit{preprocessing} techniques focus on modifying the training data in order to reduce the predictability of its sensitive attributes \cite{feldman2015certifying, calmon2017optimized}. Conversely, \textit{postprocessing} methods try to make a discriminatory predictor fair e.g. by correcting its output probabilities such that a specific fairness constraint is satisfied \cite{hardt2016equality}. However, depending on only such a post-hoc correction to achieve fairness can significantly hurt the accuracy of the predictor \cite{woodworth17a}, making it often unpractical.

DeBayes belongs to the category of \textit{algorithmic} fairness. Such methods usually try to prevent bias from affecting the learned model. A common approach is to modify the loss function, e.g. by adding fairness-inducing regularisation terms \cite{kamishima2012fairness} or by requiring that fairness constraints are met throughout the learning process \cite{zafar2017fairness, woodworth17a}. Furthermore, \cite{calders2010three} introduced naive Bayes approaches where it is assumed that latent, non-discriminatory variables can be learned that model a fair dependency between the ground truth and the data. There, a prior distribution was used to aid in finding those latent variables. Our method makes comparable assumptions, with embeddings as latent variables.

In general, the strategy of learning latent, non-discriminatory embeddings is referred to as \textit{fair representation learning}. The goal is to represent entities in the input data as informative embeddings in a metric space, without incorporating sensitive information. A popular approach is to penalize bias in the embeddings using an additional loss term. For example, this term can explicitly measure the divergence from a fairness requirement \cite{zemel2013learning} or can be the negative loss function of an adversarial classifier that attempts to predict sensitive attributes from the embeddings \cite{madras2018learning}. Of particular relevance is the Flexibly Fair Variational Autoencoder \cite{creager2019flexibly}, which enforces disentanglement between sensitive and non-sensitive information in embeddings during training. During evaluation, sensitive embedding factors can be removed, thereby avoiding their contribution in downstream tasks. The DeBayes method also attempts to represent sensitive and non-sensitive information separately during training, such that the bias can be removed afterwards. However, DeBayes differs from this prior work in that the bias is avoided through the Bayesian formulation of the objective function, rather than by reducing bias through a loss term. Consequently, bias is avoided insofar as it is present in the prior distribution, not through a manually chosen hyperparameter in the loss function. Moreover, DeBayes can incorporate any bias that can be defined through probabilities, even aggregated sensitive information.

Our interest is specifically in providing fairness for \textit{network} embedding algorithms and their link predictions. In our experiments, we use two baselines from the field of fair network embedding. The first is Fairwalk \cite{rahman2019fairwalk}: an adaptation of the popular network embedding method \textit{node2vec} \cite{grover2016node2vec}. It aims to address fairness issues by weighing the neighbourhood of each node in favour of under-represented groups. The effect is that the random walks of node2vec will be less discriminatory, with the expectation that downstream tasks will be more fair as well. The second baseline is \textit{Compositional Fairness Constraints} \cite{bose2019compositional}: an adversarial approach where the encoder is trained to produce embeddings for which its adversaries can not predict the sensitive attributes.

\section{Fairness in Link Prediction}\label{sec:fairness_in_lp}
The research in fair machine learning has given rise to several different (and sometimes conflicting) definitions of fairness. However, they are usually specialised to binary classification. In this section, we therefore extend these notions of fairness to link prediction. After introducing some notation in Section \ref{sec:notations}, we consider interpretations of fairness at the level of representation learning in Section \ref{sec:fair_repr} and the higher level of link prediction in \ref{sec:high_level}. These can then be used to inspire and evaluate the fairness properties of our method.

\subsection{Notation}\label{sec:notations}
Let $G = (V, E, A)$ be an undirected, attributed graph with set of nodes $V$, set of edges $E \subseteq V \times V$ and node attributes $A$. Without loss of generality for DeBayes (see Section~\ref{sec:generalization}), we first consider the case where every node has exactly one attribute, which is somehow sensitive, with value $s \in S$. The mapping $A: V \mapsto S$ then returns the sensitive attribute value for every node. For example, if \textit{gender} is the type of sensitive information and $v$ is a node for a male user, then $A(v) = \text{`male'}$. Below we will use $\hat{G}=(V,\hat{E},\hat{A})$ to explicitly refer to empirically observed variables.

As mentioned before, network embedding methods generally attempt to map every node $v \in V$ to a $d$-dimensional vector space $\R^d$ in a way that supports network inference tasks. Let $\bz_v \in \R^d$ be the embedding vector for node $v$. Our interest is particularly in the task of link prediction, concerned with estimating the probability $P_{u, v}$ that there exists a link $(u, v)$ between nodes $u$ and $v$.

\subsection{Fair Representations}\label{sec:fair_repr}
In fair representation learning, the aim is to find the embedding $\bz_v$ that best encodes the information about the node $v$, while its sensitive attribute value $s = A(v)$ is approximately obfuscated \cite{zemel2013learning}. The underlying idea is that if the embedding is well-obfuscated, i.e. contains no sensitive information, then those embeddings can be used in a black-box fashion to achieve fairness in downstream tasks. Building upon previous work \cite{bose2019compositional, palowitch2019monet}, we define an embedding as obfuscated if a classifier, trained on the representations and attributes of a subset of the nodes, has a low prediction accuracy on the unseen nodes. If the accuracy is equal to that of a random classifier, then the embeddings are well-obfuscated and thus the \textbf{representation bias (RB)} is removed.

For a classifier $c$, let $P_c(s, \bz_v)$ be its estimated probability (or any other score) that node $v$ with embedding $\bz_v$ has sensitive attribute value $s$. To measure its prediction performance, we chose the weighted one-versus-rest AUC score\footnote{In our experiments, we used the \texttt{roc\_auc\_score} implementation from the \texttt{scikit-learn 0.22.1} library.}. By taking the \textit{weigthed} average over the one-versus-rest AUC scores, the contribution of the set $V_s = \{v \mid A(v) = s\}$, i.e. the nodes with sensitive value $s$, is proportional to the size of their group $|V_s|$. The RB score is then defined as:
\begin{equation*}
\text{RB} = \sum_{s \in S}\frac{1}{|V_s|}\text{AUC}(\{P_c(A(v) \mid \bz_v) \mid \forall v \in V_s\}).
\end{equation*} 

While unbiased embeddings can be a useful component for a fair link prediction algorithm, limiting the fairness analysis of a link prediction method to only its generated node embeddings is not sufficient. For real-world applications, the main concern may only be for the fairness properties of the predictions themselves, not the embeddings. The specific case of CNE also presents the practical issue that it uses both embeddings and a prior to generate predictions. It may therefore still provide unfair link prediction, even when its intermediate result, the embeddings, are completely unbiased. More generally, link prediction algorithms like recommender systems may not use embeddings at all.

\subsection{High-Level Fairness Measures for Link Prediction}\label{sec:high_level}
When analysing fairness at the level of link prediction, we consider high-level measures that treat the method as a black box. To this end, we directly draw inspiration from two popular measures from the literature on fair binary classification: \textit{demographic parity} and \textit{equalized opportunity} \cite{hardt2016equality}. In addition to introducing these measures, we also consider \textit{acceptance rate parity} in our experiments.

\textbf{Demographic parity}~~ 
The definition of fairness that most closely aligns with the notion of disparate impact is \textit{demographic parity} (DP). It requires that predictions are independent of the sensitive attribute. Applied to the task of link prediction, we note that both ends may have sensitive attributes in a link. Therefore, the prediction probabilities should be independent from the attributes of both nodes in the link. As such, we define DP to be achieved when the prediction rates $\delta_{a,b}$ are equal for every pair of sensitive attribute values $a$ and $b$. Let $N_{a,b} = |\{(u, v) \mid u \in V_a, v \in V_b\}|$, i.e. all possible combinations between nodes with attribute values $a$ and $b$. The prediction rates $\delta_{a,b}$ are then computed as:
\begin{equation*}
	\delta_{a,b} = \frac{1}{N_{a,b}}\sum_{u \in V_a}\sum_{v \in V_b} P_{u, v}.
\end{equation*}
For example, an undirected social network where the sensitive attribute is $s \in \{M, F\}$ would have rates $\delta_{M,F}$, $\delta_{M,M}$ and $\delta_{F,F}$ be equal\footnote{In this undirected network, $\delta_{M,F} = \delta_{F, M}$ already holds.}. However, trying to achieve any fairness notion will usually incur a trade-off between model accuracy and fairness \cite{friedler2019comparative}. To quantify the deviation from this `perfect' DP, we define the measure as the maximum absolute difference \cite{woodworth17a} between any pair of values $\delta_{a,b}$ and $\delta_{c,d}$:
\begin{equation*}
	\text{DP} = \max_{a, b, c, d} \lvert\delta_{a,b} - \delta_{c,d}\rvert.
\end{equation*}

A crucial drawback of the concept behind demographic parity is that it may cripple the prediction quality \cite{hardt2016equality}. For example, if $M$-valued users in a social graph are more often interconnected in general, then an ideal predictor that perfectly predicts the target output, will have a higher $\delta_{M,M}$ than the other rates. In some applications, such an ideal predictor is not seen as discriminatory because it simply reflects the ground truth.

\textbf{Equalized opportunity}~~ 
A different interpretation of fairness was therefore proposed: \textit{equalized opportunity} (EO). It asserts independence between the true positive rate and the sensitive attribute. Let $N^+_{a,b} = |\{(u, v) \mid u \in V_a, v \in V_b, (u, v) \in E\}|$, i.e. all the edges $E$ of which the ends have attribute value $a$ and $b$ respectively. The true positive rate for a certain attribute pair is then $\epsilon_{a,b}$:
\begin{equation*}
	\epsilon_{a,b} = \frac{1}{N^+_{a,b}}\sum_{u \in V_a}\sum_{v \in V_b} P_{u, v} \:\mathds{1}\!\left((u, v)\in E\right).
\end{equation*}
where $\mathds{1}\!(\cdot)$ represents the identity function. Similarly to DP, we define the measure to compare EO quantitatively as:
\begin{equation*}
	\text{EO} = \max_{a, b, c, d} \lvert\epsilon_{a,b} - \epsilon_{c,d}\rvert.
\end{equation*}

\textbf{Acceptance rate parity}~~
Finally, note that other fairness measures have been introduced for link prediction. One example is a form of demographic parity based on \textit{acceptance rates} \cite{rahman2019fairwalk}, which we will also consider in our analysis. The acceptance rate $\alpha_{a,b}$ measures the relative frequency at which a link with attributes $a$ and $b$ appears in the overall top-k\% highest predictions scores. When a link between nodes $u$ and $v$ is among those top scores, it is said that $v$ is part of the recommendations $\rho(u)$ of node $u$:
\begin{equation*}
	\alpha_{a,b} = \frac{1}{N_{a,b}}|\{(u,v) \mid v \in \rho(u), u \in V_a, v \in V_b\}|.
\end{equation*}
We refer to this fairness notion as \textit{acceptance rate parity} (ARP). It is measured as the variance between rates:
\begin{equation*}
	\text{ARP} = \text{Var}(\{\alpha_{a,b} \mid a \in S, b \in S\}).
\end{equation*}

\section{The DeBayes Method}\label{sec:method}
Our goal for algorithmic fairness is to reduce the impact of the sensitive attribute on the algorithm's learning process while maintaining accuracy in its predictions. To this end, we propose DeBayes, a method that generates debiased network embeddings by modelling the sensitive information in a prior distribution.

Recall that DeBayes is based on \textit{Conditional Network Embedding} (CNE) \cite{kang2019conditional}, the principles of which will be briefly summarized in Section \ref{sec:cne}. As illustrated in Figure \ref{fig:priors_dia}, DeBayes extends CNE by utilizing a biased prior for \textit{training} and an oblivious prior during \textit{evaluation}. Their construction is discussed in Section \ref{sec:bias_prior} and Section \ref{sec:obliv_prior}. Finally, some generalizations are made in Section \ref{sec:generalization}

\subsection{Conditional Network Embedding}\label{sec:cne}

CNE finds an embedding $\bZ$ using Maximum Likelihood estimation given a model $P(\hat{G}|\bZ)$, with $\hat{G}$ the empirical graph.
Key to CNE is that the likelihood function is formulated using Bayes rule:
$P(\hat{G}|\bZ)\triangleq \frac{P(\bZ|\hat{G})\cdot P(\hat{G})}{P(\bZ)}$.
Here, the prior models properties of the graph that are hard to represent by an embedding,
such as the node degrees, or any known block structure in the graph (see below),
while the postulated form \citep{kang2019conditional} of the embedding's conditional probability $P(\bZ|\hat{G})$ assigns high probabilities to embeddings where connected nodes are embedded nearby,
and disconnected nodes are further from each other.
By modelling certain properties using the prior, the embedding itself can thus focus on representing other information.
Interestingly, $P(\hat{G}|\bZ)$ thus computed factorizes over the node pairs, such that it can immediately be used for link prediction without the need for an edge embedding or additional training.

Prior knowledge about the graph can be quantified as statistics over its structure, e.g. total density of the graph, individual node degrees, any known block structure. The prior $P(G)$ can then be constructed as the \textit{maximum entropy distribution} where the expected values of these statistics match the actual values over $G$. For example, for the degree of a particular node $u$, the prior should satisfy the following constraint where $P$ represents the prior distribution:
\begin{equation}\label{eq:degree_cne}
\sum_{v \in V} P((u, v)\in E) = \sum_{v \in V} \mathds{1}\!\left((u, v) \in \hat{E}\right).
\end{equation}
In such a prior, $P((u, v)\in E)$ is high between any node pair $v$ and $u$ if their degrees are high. Thus, high degree nodes that are connected need not be embedded very nearby in the embedding space, since their connection is already explained by the general popularity of node $u$. Conversely, if both nodes $u$ and $v$ have low node degrees, then a connection between them is a priori improbable, and they will have to be embedded closely in the embedding space.

\subsection{Training with a Biased Prior}\label{sec:bias_prior}
In DeBayes, the prior can be used to model structural information \emph{including any bias} in the networks connectivity. Specifically, we define constraints on the expected value of statistics defined in terms of the sensitive attributes of the nodes. As the prior distribution is thus dependent on the sensitive attributes $A$, we will denote it as $P(G \mid A)$ and refer to it as a \textit{biased} prior distribution. More specifically, we can enforce the constraint from Eq.~(\ref{eq:degree_cne}) within each sensitive attribute block, requiring that for every node $u$, the expected number of neighbours with a particular attribute value $s$ is equal to its empirical value. Formally, $\forall u \in V$ and $\forall s \in S$: 
\begin{equation}\label{eq:degree_bias}
\sum_{v \in V_s} P((u, v)\in E \mid A(v) = s) = \sum_{v \in V_s} \mathds{1}\!\left((u, v) \in \hat{E}\right).
\end{equation}
Consider again an example from a social graph: if user $u$ tends to mainly be connected to female users in the given graph data $G$, then this constraint will make the prior probability $P((u, v)\in E \mid A(v) = \text{`female'})$ high. At the same time, if the opposite is true for connections with male users, $P((u, v)\in E \mid A(v) = \text{`male'})$ will be low. From the perspective of node $u$, $P(G \mid A)$ is therefore biased.

Thus, DeBayes finds a debiased embedding as the Maximum Likelihood estimate $\bZ_{deb}\triangleq\arg\max_{\bZ}P(\hat{G} \mid \bZ, \hat{A})$, where
\begin{equation*}
P(\hat{G} \mid \bZ, \hat{A}) = \frac{P(\bZ \mid \hat{G})P(\hat{G} \mid \hat{A})}{P(\bZ \mid \hat{A})}.
\end{equation*}
Note that DeBayes keeps the postulated distribution of the embedding conditional on the graph $P(\bZ \mid \hat{G})$ the same as in CNE, i.e. independent of the attributes (thus $P(\bZ \mid \hat{G})=P(\bZ \mid \hat{G}, \hat{A})$). This ensures that any bias is maximally absorbed by the prior.
Taking Figure~\ref{fig:debias_example} as an example, if a political blog usually follows other blogs that are liberal-leaning, then it does not have to be embedded close to every other liberal blog. After all, it is was already known in the prior $P(G \mid A)$. Conversely, the few connections of the blog to conservative-leaning blogs will be modelled strongly in the embedding space.

\subsection{Evaluating with an Oblivious Prior}\label{sec:obliv_prior}
After fitting the embeddings with a biased prior during training, the embeddings will contain far less information about the sensitive attributes. However, the posterior edge probabilities specified by $P(G \mid \bZ_{deb}, \hat{A})$ are still computed using the biased distribution $P(G \mid \hat{A})$, and thus biased. To obtain debiased link predictions, the biased prior $P(G \mid \hat{A})$ can however simply be replaced with an \textit{oblivious} prior $P(G)$ that is oblivious to the attribute $\hat{A}$. Providing node degrees are independent of the sensitive attribute,\footnote{Of course, this is not always the case. In such situations a density prior, only constraining the overall network density, could be used instead.} a suitable candidate is a prior with constraints~(\ref{eq:degree_cne}). As this is a maximum entropy prior, sensitive attribute independence is guaranteed as long as the constraints are sensitive attribute independent. Thus we can directly use $\bZ_{deb}$ to calculate the debiased link predictions using
\begin{equation*}
P(G \mid \bZ_{deb}) = \frac{P(\bZ_{deb} \mid G)P(G)}{P(\bZ_{deb})}.
\end{equation*}
To summarize, in DeBayes, $P(\hat{G} \mid \bZ, \hat{A})$ is maximized during \textit{training} w.r.t. $\bZ$ to yield $\bZ_{deb}$, after which $P(G \mid \bZ_{deb})$ computed based on an oblivious prior can be used for debiased \textit{evaluation} (link prediction).

\subsection{Generalization}\label{sec:generalization}
In Section \ref{sec:notations}, the assumption was made that every node has exactly one attribute value $A(v) = s$ with $s \in S$. However, it is a desirable property in fair network embedding that nodes can have multiple sensitive attributes that should all be protected from discrimination \cite{bose2019compositional}. In terms of notation, each user then has a vector of attribute values $[s_0, s_1, ... \mid s_0 \in S_0, s_1 \in S_1, ...]$, where each element has its own set of possible values. 

DeBayes can easily be extended to such a case by adding the constraints from Eq.~(\ref{eq:degree_bias}) for every $S_i$. Even with overlapping constraints, it is still possible to find a maximum entropy distribution. However, one should take care that the set of constraints do not fully specify $G$, as this will make it difficult for $\bZ_{deb}$ to encode any useful information.

There is also the important case of graphs where not every attribute value is defined for every node. Such a case often occurs in recommender systems, where only the \textit{user} nodes have sensitive attributes, while the \textit{item} nodes do not. It then suffices to define an additional degree constraint for $P(G \mid A)$ that sums over those unattributed nodes. 

\section{Experiments}\label{sec:experiments}
We ran our evaluation pipeline for 10 runs, with different random seeds and train/test splits. We only report the test scores. The training set always contained approximately $80\%$ of the edges, with the test set containing the remaining $20\%$ and an equal amount of non-edges.

Evaluation was done on two datasets, detailed in Section \ref{sec:datasets}. A comparison was also made against two state-of-the-art baselines from the literature on fair network embedding as described in Section \ref{sec:related}. Their specific configuration in these experiments is described in Section \ref{sec:baselines}. The fundamental concern of our experiments was to evaluate the performance of DeBayes on the level of network embedding and link prediction. These analyses are performed in Section \ref{sec:exp_repr} and Section \ref{sec:exp_high} respectively. Finally, Section \ref{sec:discussion} wraps up with a discussion of the compared methods.

\subsection{Datasets}\label{sec:datasets}
There are two important use cases from link prediction that have been referred to throughout the paper: social graphs and recommender systems. The role of the former was filled by the \textit{\hyperref{https://www.aminer.cn/citation}{}{}{DBLP}} dataset, a co-authorship network from the DBLP database. For recommender systems, we chose the popular \textit{\hyperref{https://grouplens.org/datasets/movielens/100k/}{}{}{Movielens-100k}} dataset, a network of user-movie ratings. Both datasets are summarized in Table \ref{tab:datasets}.

\textbf{DBLP}~~
The DBLP co-authorship network \cite{Tang08KDD} is constructed from DBLP, a computer science bibliography database. In our analysis, we only considered publications at \textit{KDD}, \textit{ECML-PKDD}, \textit{DAMI}, \textit{NIPS}, \textit{JMLR}, \textit{ICML}, \textit{MLJ} and \textit{ICLR}. From the authors of papers accepted at these venues, we chose those for which the country of their affiliation could successfully be parsed from the 'org' field. The corresponding continents\footnote{The country/continent mapping as defined by the \texttt{pycountry-convert} package.} are their sensitive attribute. Due to strong under-representation in the data we did not consider the continents of Africa and Antartica in our analysis. The authors were linked if they had collaborated at least once. Disconnected authors were removed.

\textbf{Movielens-100k}~~ 
The Movielens-100k (ML-100k) dataset is a staple in recommender systems research due to its manageable size and rich data. The nodes represent users or movies and only user-movie links exist in the network. The 5-star ratings were not included in our experiments. All user attributes are seen as sensitive: gender, age and occupation. Ages were binned into 7 age groups in order to make the attribute values categorical.

\begin{table}[tb]
\caption{Dataset statistics. For every set of possible attribute values $S$ their cardinality is given by $|S|$. The dataset names link to the URL where they can be downloaded.}
\label{tab:datasets}
\vskip 0.1in
\begin{center} 
\begin{small}
\begin{tabular}{lcccc}
	\toprule
	\sc Datasets & $|V|$ & $|E|$ & $S$ & $|S|$ \\
	\midrule
	\sc \hyperref{https://www.aminer.cn/citation}{}{}{DBLP} & 3,980 & 6,965 & continent & 5 \\ \hline
	\multirow{3}{*}{\sc \hyperref{https://grouplens.org/datasets/movielens/100k/}{}{}{ML-100k}} & & \multirow{3}{*}{100,000} & gender & 2 \\
	& 2,625  &  & age bracket& 7\\
	&  &  & occupation   & 21 \\
 	\bottomrule
\end{tabular}
\end{small}
\end{center}
\vskip -0.2in
\end{table}

\begin{figure*}[tbh]
	\centering
	\begin{subfigure}[tbh]{\textwidth}
		\centering
		\includegraphics[width=\textwidth]{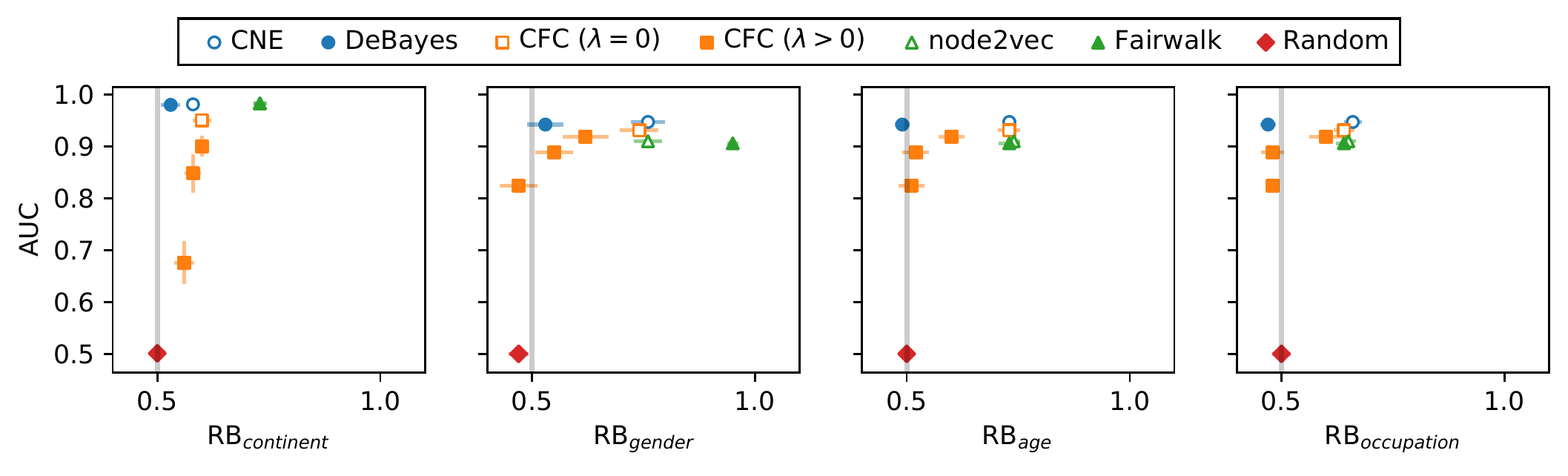}
		\caption{The RB when using a \textit{logistic regression} classifier.}
	\end{subfigure}
	\hfill
	\begin{subfigure}[tbh]{\textwidth}
		\centering
		\includegraphics[width=\textwidth]{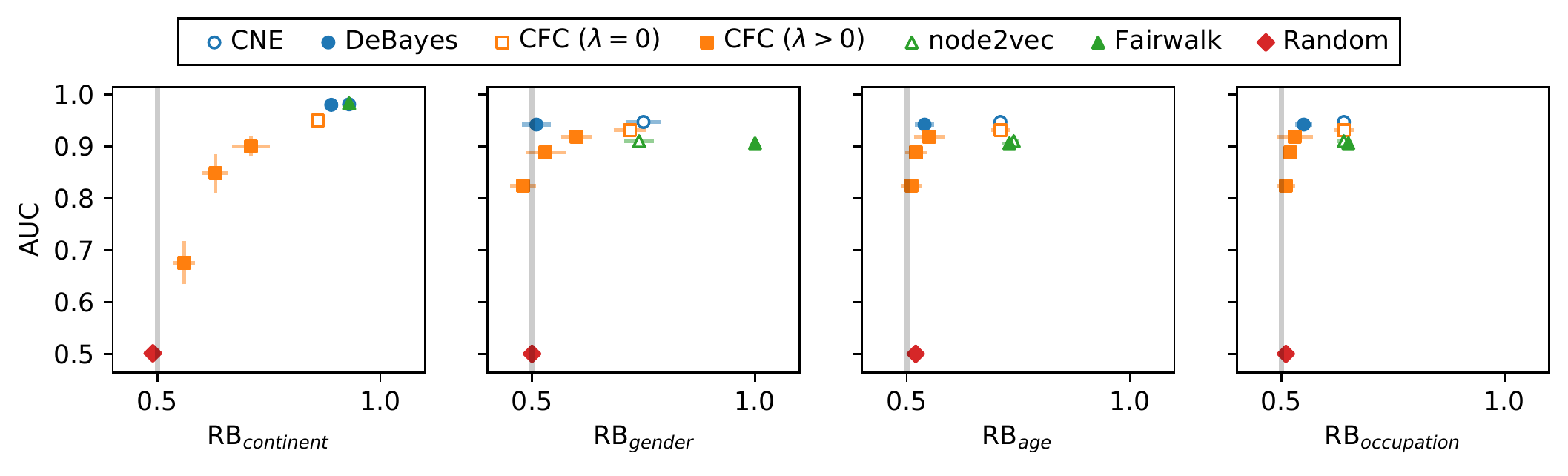}
		\caption{The RB when using an \textit{SVM} classifier with \textit{RBF} kernel.}
	\end{subfigure}
	\vskip -0.1in
	\caption{For various configurations of every method, the trade-off between link prediction AUC and RB when measured with classifiers of different strengths. Method configurations without any fairness adaptations, e.g. no biased training prior for DeBayes, were given \textbf{empty markers}. The ideal value for AUC is $1$ and the ideal value for RB is $0.5$ (indicated by the grey vertical line). The plotted values are the mean test set values over 10 random seeds. Symmetric error bars show standard deviation in both directions. Note that RB$_{continent}$ was evaluated on the DBLP dataset, while the other RB values result from experiments on ML-100k.}
	\label{fig:repr_bias}
	\vskip -0.05in
\end{figure*}

\subsection{Configurations}\label{sec:baselines}
Our proposed method is DeBayes. As baselines, \textit{Compositional Fairness Constraints} (CFC) \cite{bose2019compositional}, \textit{Fairwalk} \cite{rahman2019fairwalk} and a uniformly \textit{Random} predictor were used. Hyperparameter optimization was performed on the most influential parameters with link prediction AUC as a target. For that optimization, 20\% of the training set was used as a validation set. Other parameters were left at their default values.

\textbf{DeBayes}~~
The DeBayes method was trained using both an oblivious training prior and a biased training prior. Evaluation was always done using an oblivious training prior. Recall that DeBayes with an oblivious training and evaluation prior is equivalent to CNE. For dimensionality, $d=8$ was chosen out of $d \in \{8, 16\}$. The parameter $\sigma_2$ was kept constant, and $\sigma_1=0.7$ was chosen from $[0.4, 0.9]$.

\textbf{Compositional Fairness Constraints (CFC)}~~
We used the dot-product encoder for CFC in both datasets. The strength of the adversarial term in the loss function is specified by the parameter $\lambda$. We trained models with $\lambda \in \left[0, 5, 25, 100\right]$. The maximum value was $\lambda = 100$, as that already resulted in a much lower prediction performance than the other algorithms. The learning rate parameter was evaluated over the range $[0.0001, 0.01]$. For DBLP, a learning rate of $0.01$ was used; for ML-100k, it was left at its default value of $0.001$.

\textbf{Fairwalk}~~
Since Fairwalk can not be trivially extended to multiple sensitive attributes, it was only given `gender' as a sensitive attribute for the ML-100k dataset. We present results for Fairwalk with and without the fair random walk modification. Note, the latter is equivalent to the original node2vec. The default dimensionality of $d=128$ was used.

\subsection{Analysis of Representation Bias}\label{sec:exp_repr}
The representation bias (RB) (Section \ref{sec:fair_repr}), is measured in our experiments using two classifiers: a logistic regression classifier and an SVM with a non-linear kernel. Their hyperparameters were tuned separately on every algorithm's set of node embeddings and for every attribute. After tuning, each classifier is trained on $80\%$ of the nodes and then tested on the remaining embeddings.

In Figure \ref{fig:repr_bias}, the trade-off is shown between link prediction AUC and RB. By using a biased prior, DeBayes clearly produces significantly debiased embeddings with respect to all the ML-100k attributes (gender, age and occupation), since both classifiers can only achieve random-like performance in predicting them. At the same time, the decrease in AUC is limited. For DBLP, the logistic regression classifier becomes less successful at estimating the continent, but the non-linear SVM can still make a strong separation.

The embeddings from CFC's adversarial approach also appear to be well-debiased, even for the seemingly difficult case of reducing $RB_{continent}$ against a non-linear SVM classifier. However, a significant decrease in link prediction AUC is incurred at the same time. Furthermore, it was observed that on $RB_{gender}$, the bias in the embedding surprisingly \textit{increased} using the fair random walk. In contrast to the other algorithms, both its AUC and $RB_{gender}$ score are worse for the configuration with fair random walks. Recall that on ML-100k, only the `gender' attribute was marked as sensitive for Fairwalk, explaining why measures for `age' and `occupation' remained relatively constant.

\subsection{Analysis of High-Level Fairness}\label{sec:exp_high}
\begin{figure}[tbh]
	\begin{center}
		\centerline{\includegraphics[width=\columnwidth]{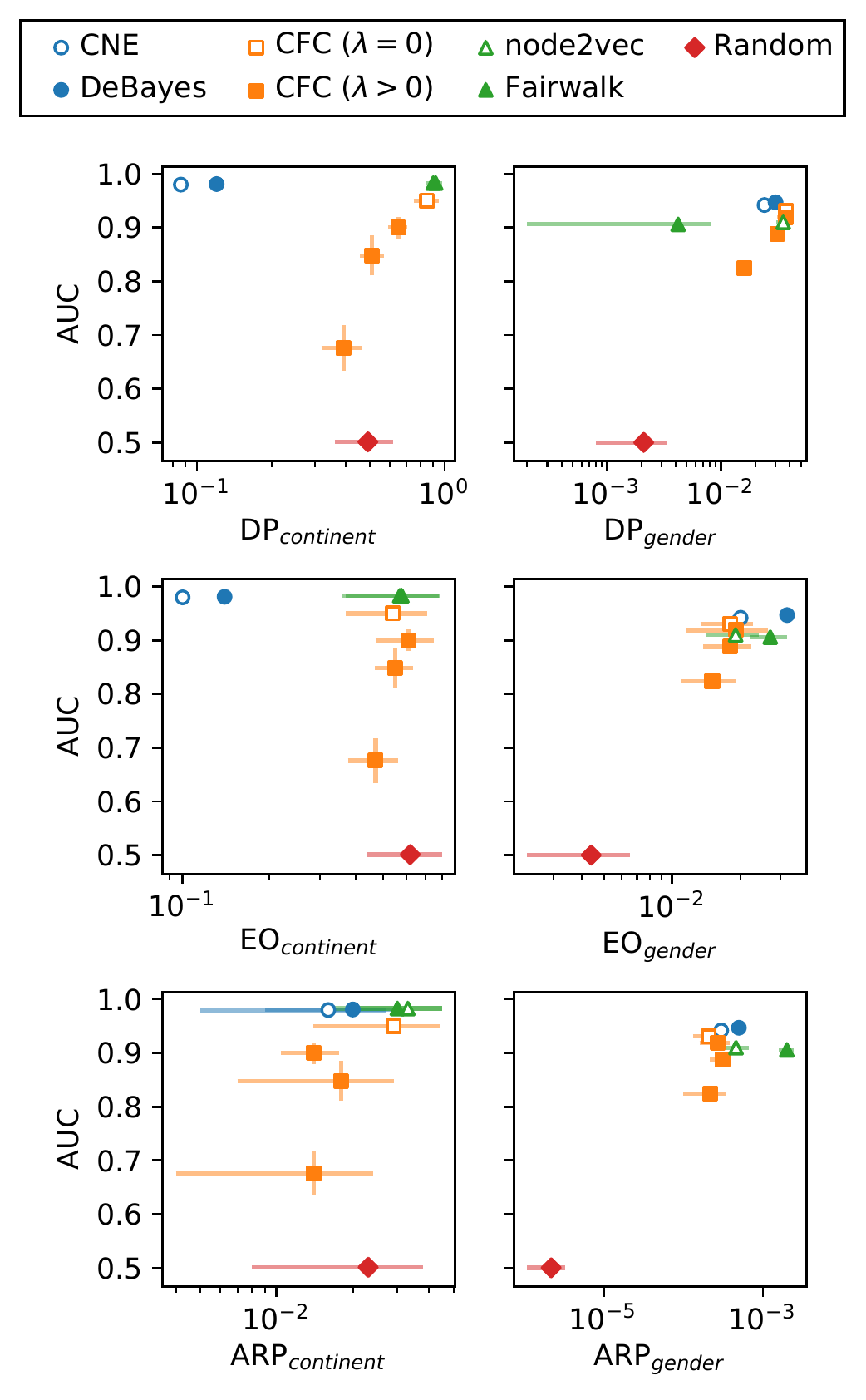}}
		\caption{For various configurations of every method, the trade-off between link prediction AUC and high-level fairness measures. For ML-100k, the measures were only computed for the 'gender' attribute. In contrast to Figure \ref{fig:repr_bias}, the ideal value for all high-level measures is $0$. Also, each plot has a different, logarithmic x-axis.}
		\label{fig:high_level}
	\end{center}
	\vskip -0.3in
\end{figure}

Results on the high-level measures are illustrated in Figure \ref{fig:high_level}. On DBLP, the oblivious prior configuration for DeBayes already scores the best on the traditional fairness measures: DP and EO. Yet, when trained with a biased prior, the predictions appear even more fair. On the other measures, DeBayes also improves upon its unfair configuration.

Although CFC is also successful at improving on all fairness measures given a large enough $\lambda$ value, it must do so with a sharply dropping AUC. Furthermore, Fairwalk maintains its strong AUC on both datasets, but with only a significant improvement on $DP_{gender}$. On the other hand, Fairwalk scores worse on $EO_{gender}$ and $ARP_{gender}$.

Interestingly, while both DeBayes and CFC were able to reduce $RB_{gender}$ to random-like values, they are unable to come close to the random predictor's high-level fairness. However, the achieved fairness measures were already quite low without any fairness adaptations.

\subsection{Discussion}\label{sec:discussion}
Both DeBayes and CFC are able to significantly debias the node representations, though they depend on different strategies. In the case of DeBayes, the learning process is allowed to freely minimize its loss function. As long as the prior is powerful enough to model the bias in the data well, even a non-linear classifier is unable to predict the sensitive information. The DeBayes method also generates more fair predictions by using an oblivious evaluation prior on top of its debiased embeddings. Overall, the results indicate that DeBayes indeed focusses on taking out biased information, while still learning informative embeddings.

This is in contrast with CFC, where an adversarial term causes the embeddings and downstream tasks to be more fair. The issue is that such a term could be unstable, complicating the learning of non-discriminatory dependencies between the ground truth and the data; a hypothesis that is supported by a sharp drop in AUC scores for comparatively mild fairness improvements. On the other hand, Fairwalk appears to make its embeddings more biased towards the sensitive attribute on ML-100k. It was only successful at lowering its DP, though it barely saw its AUC affected. Moreover, the fair random walk strategy did not have a significant impact on the DBLP dataset. One possible cause is the sparsity of the network, causing there to be many neighbourhoods where nodes are only connected with other nodes that have the same attribute value. In such neighbourhoods, Fairwalk does not change the random walk compared to node2vec.

In addition to prediction performance and fairness, there are other useful properties for an algorithmic fairness method, such as \textit{scalability}. Recall that for DeBayes, the prior distribution is computed completely before learning the embeddings. The latter step is equivalent to the original CNE method, which was shown to be faster than state-of-the-art network embeddings like node2vec on a single process/thread for large datasets \cite{kang2019conditional}. Furthermore, for the optimization problem of finding the prior distribution, DeBayes only adds constraints that are at most as complex as those in (\ref{eq:degree_cne}). They only increase the computational complexity by a constant factor and do not introduce additional memory requirements. However, we have so far only considered sensitive attributes with categorical values. If we had not performed binning of age groups for ML-100k, then the number of constraints (one for every age) would have become intractable. In order to take more complex sensitive attributes into account, such as those with continuous values, different constraints would have to be defined.

\section{Conclusion}\label{sec:conclusion}
DeBayes is a conceptually simple and elegant method for finding debiased network embeddings based on Conditional Network Embedding.
DeBayes results in substantially debiased node embeddings in almost all cases, and the link predictions score well on three important high-level fairness measures. Even though there is inevitably a trade-off between fairness and prediction AUC, the cost in prediction quality by using DeBayes is limited. The method is also flexible enough to incorporate multiple types of sensitive attributes and does not depend on minimizing a fairness-related loss term. Instead, it depends on the ease of modelling the sensitive information through a prior distribution.

Further research into the training prior could allow for more kinds of sensitive attributes, e.g. with continuous values. Moreover, the choice of the evaluation prior is flexible and could be modified to be even more debiased (e.g. when dependencies exist between degrees and the sensitive attribute).
Finally, principles underlying DeBayes could be generalized and applied to other areas in fair machine learning.

\section*{Acknowledgements}
The research leading to these results has received funding from the European Research Council under the European Union's Seventh Framework Programme (FP7/2007-2013) / ERC Grant Agreement no. 615517, from the Flemish Government under the ``Onderzoeksprogramma Artificiële Intelligentie (AI) Vlaanderen'' programme, and from the FWO (project no. G091017N, G0F9816N, 3G042220).

\newpage
\bibliography{references}
\bibliographystyle{icml2020}

\end{document}